%% file: main.tex
\pdfoutput=1

\documentclass[11pt]{article}

\usepackage[final]{acl}
\usepackage{times}
\usepackage{latexsym}
\usepackage[T1]{fontenc}
\usepackage[utf8]{inputenc}
\usepackage{microtype}
\usepackage{inconsolata}
\usepackage{graphicx}
\usepackage{amsmath}
\usepackage{amssymb}
\usepackage{tcolorbox}
\usepackage{listings}

\usepackage{subcaption}
\usepackage{paralist}
\usepackage{float}
\usepackage{comment}
\usepackage[export]{adjustbox}
\usepackage{algorithm}
\usepackage{algorithmic}
\newcommand{\ie}{i.e.,\ }
\usepackage{booktabs}
\usepackage{multirow}
\usepackage{array}
\usepackage{rotating}
\usepackage{fp}
\usepackage{makecell} 
\usepackage{tabularx}
\usepackage{listings}
\usepackage{soul}  
\usepackage[textsize=small]{todonotes}

\lstdefinestyle{cppstyle}{
  language=C++,
  basicstyle=\ttfamily\footnotesize,
  numbers=left,
  numberstyle=\tiny,
  stepnumber=1,
  numbersep=5pt,
  showstringspaces=false,
  tabsize=2,
  breaklines=true,
  breakatwhitespace=false,
  escapeinside={\%*}{*)},
  frame=single,
  xleftmargin=2pt,
  xrightmargin=2pt
}

\newcommand{\calcpercent}[2]{%
    \FPeval{\result}{round(#1/#2*100, 1)}%
    \result\%%
}

\usepackage{listings} 

\newcommand{\eat}[1]{}

\title{Steered Generation via Gradient Descent on Sparse Features}

\author{Sumanta Bhattacharyya \\
  Department of Computer Science \\
  University of Illinois Chicago \\
  \texttt{sbhatt54@uic.edu} \\\And
  Pedram Rooshenas \\
  Department of Computer Science \\
  University of Illinois Chicago \\
  \texttt{pedram@uic.edu} \\}

\newcounter{inlineitem}
\begin{document}
\maketitle 
\begin{abstract}
Large language models (LLMs) encode a diverse range of linguistic features within their latent representations, which can be harnessed to steer their output toward specific target characteristics. In this paper, we modify the internal structure of LLMs by training sparse autoencoders to learn a sparse representation of the query embedding, allowing precise control over the model’s attention distribution. We demonstrate that manipulating this sparse representation effectively transforms the output toward different stylistic and cognitive targets. Specifically, in an educational setting, we show that the cognitive complexity of LLM-generated feedback can be systematically adjusted by modifying the encoded query representation at a specific layer. To achieve this, we guide the learned sparse embedding toward the representation of samples from the desired cognitive complexity level, using gradient-based optimization in the latent space.
\end{abstract}
\section{Introduction}
\input{intro_new.tex}

\section{Method}
\input{method.tex}

\section {Experimental Setup} 
\input{experiment.tex}

\input{evaluate.tex}

\section{Related works}
\input{bg.tex}

\section{Conclusion and Future Work}

Our analysis of feature space sparsity in cognitive transformation demonstrates that training independent SAEs for each query attention head from mid layer can maintain separate characteristics for each cognitive style. While sparse features typically correspond to a specific style, some may also be activated by closely related styles, as SAEs likely face the challenge of polysemanticity where activations respond to multiple meanings and concepts. This could explain the observed gap between SAE-generated steered text and target text. Although mono-semanticity (where a single neuron corresponds to single feature) is difficult to achieve due to architectural or computational constraints, controlling the degree of polysemanticity remains an important direction for improving precision.
Since we currently train SAEs from the internal activations of fine-tuned LLM, our approach is scalable across different LLM sizes. Future research directions include investigating the co-training of SAEs with LLM to achieve native interpretability through joint optimization.

\bibliography{custom}
\bibliographystyle{acl}

\clearpage

\input{appendix}

\end{document}

%% file: intro_new.tex
\eat{this is not an interpretability paper, the introduction must start with generation}

LLMs have emerged as the de facto natural language interface for interacting with AI systems. In most modern applications, prompt engineering is crucial for steering the output generation process, controlling the nuance and style of text, and tailoring it to suit the audience's needs -- including factors like empathy and impartiality. 

Research on controlled text generation in large language models (LLMs) has been extensively explored for general-purpose applications (see the comprehensive survey by~\citet{ctg-survey}). Unlike standard text generation, controlled text generation~\cite{mireshghallah2022mix} guides LLMs to generate outputs that adhere to specific constraints.

\eat{With the advent of large pre-trained models, in-context learning (ICL)~\cite{brown2020language} has emerged as a significant paradigm for controlled text generation. It enables LLMs to adapt their behavior based on demonstrations or instructions embedded directly within the input prompt. While ICL is valuable for leveraging the inherent capabilities of pre-trained models, it has limitations.} Controlling text generation solely through prompting can be ineffective~\cite{ye2022unreliability}, as prompts often require extensive contextual information to generate relevant responses. When additional instructions for controlling style and nuances are included, these details may be overlooked or lost within the large prompt context~\cite{liu2023lost}. Moreover, open-source and smaller LLMs provide limited control over generated text when relying solely on prompt-based methods~\cite{salinas2024butterfly}.

For instance, a model to understand and execute instructions like "generate the response in [X] style", it needs extensive exposure to such instructions during training to reliably control output style via its internal sentence representation. Models like GPT-4 ~\cite{achiam2023gpt} and Claude 3.5 Sonnet generally follow style instructions effectively, whereas smaller models, such as Phi-3 ~\cite{abdin2024phi} (trained through knowledge distillation from larger models like GPT-4), often struggle to adhere precisely to specified style constraints. 

One approach to mitigating this in smaller language models is to incorporate demonstrations in the prompt, a technique known as in-context learning~\cite{{brown2020language}}. We observe that Phi-3's ability to follow style-specific instructions improves as more demonstration are provided. However, one caveat is that increasing the number of demonstrations leads to significant memory consumption and computational overhead, particularly with the larger context windows needed for complex tasks.

Latent space-based steering methods, often referred to as activation engineering, offer an effective approach for controlled text generation by strategically intervening in the activations of LLMs~\cite{Dathathri_Madotto_Lan_Hung_Frank_Molino_Yosinski_Liu_2019,Liu_Sap_Lu_Swayamdipta_Bhagavatula_Smith_Choi_2021,Khalifa_ElSahar_Dymetman_2020}. These interventions are achieved by applying steering vectors -- directional components in the activation space that correspond to specific semantic or stylistic features -- during inference. The steering vectors shift the model's behavior in predictable ways~\cite{hernandez2023inspecting,sun2024massive}. Unlike in-context learning, activation engineering allows for control over multiple independent attributes (e.g. style, tone, or formality) by disentangling these features in the activation space. However, the effectiveness of this method relies on the interpretability of the activation features being manipulated. Recent advancements suggest that introducing sparsity in activation functions -- via methods such as Sparse Auto-Encoders (SAE)~\cite{Makhzani2013kSparseA} -- can significantly enhance interpretability~\cite{cunningham2023sparse,bricken2023towards}. Increased sparsity improves disentanglement in the latent space, enabling more precise and consistent control over independent attributes during text generation.

\par

In this paper, we propose a novel approach for controlled text generation using sparse feature-based steering during inference. Inspired by prototypical networks~\cite{snell2017prototypical}, originally designed for few-shot learning, we represent each attribute class of interest as a prototype distribution in a latent space defined by SAE and steer the encoded features to increase the likelihood of output for the desired prototype.

\eat{we enable efficient classification and systematic steering of features to achieve desired text attributes. This approach naturally extends to multiple attributes, offering flexible and nuanced control over text generation.}
\begin{figure*}[htbp]
\centering
  \includegraphics[width=1.08\linewidth]{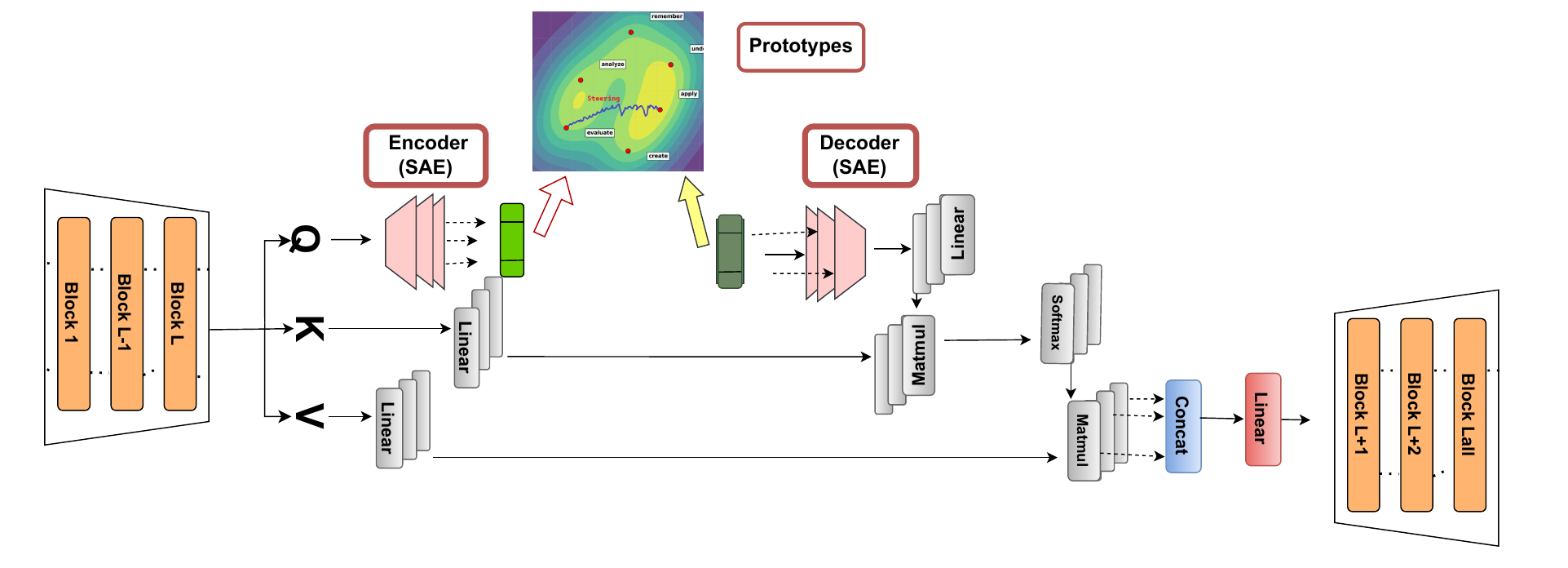}
  \caption{Training architecture of our method. We divide the LLM  into Enc-LLM, Upto Layer L till the sparse encoders and Dec-LLM for the rest of the layers including the sparse decoders.
}
  \label{fig:entire}
\end{figure*}

We evaluate our method on synthetically generated dataset, specifically designed to capture varying classes of cognitive style for the given input. Our experiments demonstrate that SAEs effectively capture and distinguish the nuances between these classes, and further investigation reveals their utility in steering generation toward the desired style. Our primary contributions and findings are the following:

\noindent\textbf{Novel Dataset Creation}: We develop a novel dataset by employing proprietary LLMs (Claude) to generate code review problems accompanied by feedback of varying cognitive styles (Bloom's taxonomy)~\cite{uugur2015self}. This dataset facilitates the examination of the steering across nuanced styles of text. 

    
    \noindent \textbf{Enhanced Interpretability via Attention Mechanisms}: Our observation reveals that query representations within attention heads are particularly effective in capturing distinguishing textual features, highlighting the critical role of attention components in processing and interpreting input data.

    \noindent \textbf{Text Generation Steering}: We successfully control the generation of texts to achieve specific styles by applying a gradient-based approach to sparsely encoded features, validating the practical applicability of our prototype-based steering method.

    \noindent \textbf{Evaluation}: We conduct extensive evaluations across various layers and dimensions to demonstrate our approach's effectiveness.

%% file: method.tex
Intuitively, we aim to steer text generation from one nuanced style to another by modifying the model’s internal attention distribution --specifically, by adjusting the query embedding toward a target vector.  At a high level, attention layers direct the vector representation of tokens to capture contextual information. Each attention layer achieves this by creating multiple distributions over the embeddings of other tokens in the context, adjusting each token's embedding to reflect its surrounding context. By modulating these distributions, we gain control over text generation. In particular, each attention distribution is generated via a dot product between two representations of tokens: the query and the key. However, we observed that the query embedding often captures a mixture of multiple target styles, reducing its effectiveness. To address this, we employ a sparse representation obtained via SAE, which enhances the distinctiveness of features associated with each style. By transitioning between these sparse representations, we achieve real-time control over the nuanced style of the generated output during inference. Figure~\ref{fig:entire} provides a schematic over of the proposed method.

\subsection{Sparse Encoding of Nuanced Styles}
Sparse autoencoder (SAE)~\cite{konda2014zero,lee2007sparse} is a type of autoencoder designed to learn efficient data representations by enforcing a sparsity constraint, often through $L_1$ regularization. This constraint encourages the model to activate only a small subset of neurons at any given time, promoting the discovery of distinct and meaningful features. Unlike standard autoencoders, which often learn dense low-dimensional representations, SAEs specialize in high-dimensional sparse representations. These representations are particularly useful for feature disentanglement and enhancing interpretability, as they encourage the network to encode independent and interpretable factors of variation in the data. 
It proves to be effective for analyzing and interpreting latent representations in large language models (LLMs)~\cite{cunningham2023sparse,bloom2024open,marks2023some}. \\
By learning a sparse representation for the query vector with SAEs, we can learn a disentangled representation for every nuanced style of text, which facilitates the transition from one style to another in the latent space.

To effectively represent the styles in the latent space trained by SAE, we borrow the idea from Prototypical Networks~\cite{snell2017prototypical} introduced for few-shot learning.
We define class prototypes as an embedding function 
\( \text{Enc-LLM}{\phi}: \mathbb{R}^{S\times D} \to \mathbb{R}^{S\times A \times H} \) where $S$ is the maximum sequence length (including both prompt and response), $D$ is dimension of token embedding, and $A$ is the number of attention head for intervened layer, $H$ is the size latent representation in SAE, and $\phi$  are learnable parameters of SAE.

\subsection{Steering Output Style} For each style class \(k\), we define a support set $D_k$ consisting of training examples labeled as style class \(k\): \( D_k = \{((x_i, y_i), l_i) : l_i = k\} \), where $x_i$ is prompt, $y_i$ is response, and $l_i$ is style class label. For each class, we define the class prototype $c_k$ as the mean embedding of the support examples:
$c_k = \frac{1}{|S_k|} \sum_{(x_i, y_i, l_i) \in S_k} \text{Enc-LLM}_{\phi}(x_i,y_i)$. 
For each class $k$ we define the probability distribution for each example $(x,y)$ using a softmax over negative distances to the corresponding prototype class, $c_k$: $p_{\phi}(l = k | x, y) = \frac{\exp(-d(\text{Enc-LLM}_{\phi}(x,y), c_k))}{\sum_{k'} \exp(-d(\text{Enc-LLM}_{\phi}(x,y), c_{k'}))}$.

To steer the generation, we modify the embedding of $\text{Enc-LLM}(x,y)$ to increase the target class log-probability, $\log p_\phi( l = t \mid x,y)$.
Instead of directly using the center embedding of the target class ($c_T$) for decoding, which maximizes the defined probability distribution, we use gradient descent to increase the probability. A direct assignment to $c_T$ causes a loss of variability in output styles (for additional study please refer to section {\ref{sec:tcd}}). By optimizing via GD, we allow for a more flexible transformation that respects the latent structure while encouraging stylistic alignment. Moreover, GD allows a gradual and interpretable shift in the latent space, preserving contextual consistency while moving toward the desired style. 
After reaching the desired embedding $z$, we use it to reconstruct the query embedding and decode the output. Algorithm~\ref{alg:steering} illustrates the steering process. 



\begin{algorithm}
\caption{Class-based Steering using Sparse Autoencoder}\label{alg:steering}
\begin{algorithmic}[1]
\STATE \textbf{Input:} Source text $x$, target class $c_T$,  $D_{\text{train}}$, LLM, Enc-LLM and Dec-LLM. 
\STATE \textbf{Output:} Steered text toward target class

\STATE $C \gets$ All classes in $D_{\text{train}}$
\FOR{$c \in [C]$}  
    \STATE $S_c \gets \{x' \mid (x',y') \in D_{\text{train}} \wedge y' = c\}$
    \STATE $\mu_c \gets \frac{1}{|S_c|} \sum_{x'\in S_c} \text{Enc-LLM}(x')$
\ENDFOR
\STATE $y \gets \text{LLM}(x)$
\STATE $z \gets \text{Enc-LLM}(x,y)$
\WHILE{$\|\nabla_z \log P_{\phi}(y = c_2| z) \|^2_2 > \epsilon$ }
    \FOR{$c \in [C]$} \STATE $d_c \gets \|z - \mu_c\|_2$ \ENDFOR
    \STATE $p_{\phi}(y = c_T | z) \gets \frac{\exp(-d_{c_T})}{\sum_{k=1}^C \exp(-d_k)}$ 
    \STATE $z = z + \eta\frac{\partial}{\partial z} \log P_{\phi}(y = c_2| z)$   
\ENDWHILE
\STATE \textbf{return} $\text{Dec-LLM}(x,z)$

\end{algorithmic}
\end{algorithm}

\subsection{Training Sparse Autoenoders}
Our SAE architecture is a single layer ReLU autoencoder (similar to~\citet{bricken2023towards}).
For the input embedding $q_{in}$, encoder $W_e$ and decoder $W_d$, the latent embedding of SAE is defined as 
$z_{sae} = \text{ReLU}(W_e q_{in} + b_e)$.

The loss function combines the reconstruction loss, sparsity constraints, as well as a bias decay term:

\begin{align}
\mathcal{L}(x) = \underbrace{\|\hat{q_{in}} - q_{in}\|_2^2}_{\text{Reconstruction loss}} + \underbrace{\alpha\|z_{sae}\|_1}_{\text{Sparsity loss}} + \underbrace{\beta\|b_e\|_2}_{\text{Bias decay}},
\end{align}
where $\hat{q_{in}} = \frac{W_d}{\max(\|W_d\|_2, \epsilon)}z_{sae}$,
 $\alpha$ is the regularization strength of $L_{1}$, $\beta$ is the bias decay parameter.
We train it on the activations of query attention heads from self attention layer. Since each attention head operates independently in a multi-head attention mechanism to capture different aspects of the input data, we need to train multiple SAEs independently each for the query attention heads. Learned sparse representations by each SAEs provide interpretable continuous features for desired attributes in the data. Hereby steering the generation process toward outputs with specified characteristics and maintaining high fluency~\cite{liang2024controllable}. We observe that the trained SAEs on the earlier or later layers of the LLM are not as beneficial as compared to the middle layers (such as layer 15). Based on the position of the SAE at particular layer L ($L_{all}$ being the number of layers), We define Enc-LLM as the collection of layers till the sparse encoders at L, rest ($L_{all}$-L) layers including the sparse decoders are defined as Dec-LLM. 

\eat{
To quantify the benefit of these sparse features, we follow a similar~\cite{snell2017prototypical} few-shot classification. We collected fixed number of $x'$ from training data for each class $c \in [C]$ as support examples and take the mean of the sparse features of these support examples to create class-specific centers (prototypes) ($\mu_c$). For an unseen example, association with a particular prototype is determined by its proximity to the prototype of each class. This accuracy defines how well the prototypes have captured the distinctness between the different class feature. For an quantitative analysis, Please refer to Table \ref{tab:matching-analysis} or refer to the Figure \ref{fig:combined_analysis} for the visual plot.

}




This gradient is then used to guide the steering process toward the desired class:
\begin{align}
z_{i \rightarrow j} = z_{i} + \eta\frac{\partial}{\partial z} & \log P_{\phi}(y = j|c, z)\label{eq:steering}
\end{align}
Here, $\eta$ is a hyperparameter that controls the step size of the gradient update.

%% file: experiment.tex
We evaluate our approach on feedback generation in educational settings, focusing specifically on cognitive complexity transformation where feedback is adjusted according to Bloom's taxonomy~\cite{blooms}. Due to data scarcity, we generate a synthetic dataset using proprietary LLM ( Claude 3.5 Sonnet (v1)) through prompt engineering, a commonly adopted practice in this domain (see~\cite{liu2023pre,lu2023machine} for a comprehensive
survey). 

We also employ proprietary LLM (Claude 3.5 Sonnet (v1)) as a reference evaluator to categorize the post-steering generated feedback according to Bloom's taxonomy cognitive levels.

\eat{The effectiveness of our steering approach is visualized through stacked bar charts, where each bar illustrates how the steered generations are distributed across different cognitive levels in the dataset.}

\subsection{Dataset}

We construct a novel dataset to evaluate nuanced style transformation. The dataset targets ten fundamental C++ programming concepts (recursive functions, arrays, call-by-reference, switch statements, do-while loops, for-loops, while-loops, pointers, strings, and vectors), each represented by multiple coding questions and systematic feedback variations. At the initial stage, we prompt the LLM to generate 20 questions spanning easy, medium, and hard difficulties, each with its correct answer. Then, we have the LLM create four variations of each answer that contain deliberate errors. Finally, we generate feedback for both the original correct answer and all variants, using Bloom's taxonomy levels~\cite{blooms} (Remember, Understand, Apply, Analyze, Evaluate, and Create). \eat{For detailed strategy, please refer to section \ref{sec: datacretae} }


\begin{figure}[!t]
    \centering

\begin{lstlisting}

Below is an instruction that describes a task, paired with an input that provides further context. Write a response that appropriately completes the request.}

### Instruction:
Provide feedback on the following code
Write a program to swap two numbers using a function with call by reference.

### Code:

#include <iostream>
using namespace std;

void swap(int &a, int &b) {
    a = b;
    b = a;
}

int main() {
    int x = 5, y = 10;
    cout << "Before swap: x = " << x << ", y = " << y << endl;
    swap(x, y);
    cout << "After swap: x = " << x << ", y = " << y << endl;
    return 0;
}
### Response:

\end{lstlisting}

\begin{lstlisting}
The code structure reveals a logical flow issue where sequential assignments cause data loss. The relationship between the swap function and main program is correct, but the internal swap mechanism fails to maintain value integrity during the exchange process.

\end{lstlisting}

    \caption{A sample data point with Alpaca template.}
    \label{fig:sample-data}
\end{figure}

\par

\eat{For cognitive complexity transformation, we guided the LLM to produce responses aligned with Bloom’s Taxonomy into six hierarchical levels: Remembering, Understanding, Applying, Analyzing, Evaluating, and Creating, progressing from lower-order thinking skills to higher-order thinking skills, reflecting a range of pedagogical depth.}


Our dataset comprises 3,000 data points, calculated as follows: 10 topics × 20 questions × 3 difficulty levels × 5 code variants (one correct, four with errors). We split the data by topic into training (50\%), validation (20\%), and test (30\%) sets to ensure conceptual separation between splits. 

\eat{
\subsection{Variations in verbosity}
To create length-varied feedback, we employed a three-stage prompting pipeline using LLMs: 
\begin{inparaenum}[(1)]
\item \textbf{Base Questions:} LLM, acting as a C++ expert, generated questions spanning ten function types and three difficulty levels (easy, medium, hard).
\item \textbf{Code Variations:} LLM, assuming the role of a student, produced four error-prone code variations for each question.
\item \textbf{Feedback Generation:} LLM, as a code review assistant, provided feedback at three length granularities (short,medium, long) for both the original and the modified codes.
\end{inparaenum}

\subsection{Feedback Cognitive Variations}
For cognitive-style feedback, we guided an LLM configured as a computer science instructor to produce responses aligned with Bloom’s Taxonomy~\cite{uugur2015self}. Bloom's Taxonomy classifies educational learning objectives into six hierarchical levels: Remembering, Understanding, Applying, Analyzing, Evaluating, and Creating, progressing from lower-order thinking skills to higher-order thinking skills, reflecting a range of pedagogical depth.

For all future reference, we will refer to these two datasets as  "style" data.
}

\subsection{Models and Configurations} 
In all of our experiments, we primarily use the \emph{phi-3-mini-4K instruct} model~\cite{abdin2024phi} as the base model, except where otherwise noted. While prompting the input into the model (Figure~\ref{fig:sample-data} shows input data sample), we adapt the Alpaca~\cite{alpaca} format, with maximum context size of 512 tokens. We perform one epoch of fine-tuning on the training split of our dataset. Subsequently, we use the query activations from this fine-tuned model to train the SAEs.

We train multiple independent SAEs on each query attention head of a single layer of the fine tuned base model. Each SAE corresponds to one query attention head with the query embedding size of 96 and maps it to a high-dimensional latent space of 512. For training, we use the $L_{1}$ coefficient ( \(\lambda\)) values of $3\times10^{-4}$, $3\times10^{-3}$, and $3\times10^{-2}$ for different settings.
We train the SAEs for 40 epochs till it converge. We employ a linear warm-up followed by a cosine decay scheduler to dynamically adjust the learning rate, starting from \(3 \times 10^{-5}\). Optimization is performed using the Adam optimizer with \(\beta_{1} = 0.90\) and \(\beta_{2} = 0.99\).

We evaluate our method using the sparse features ($z$) from layer 15 with a \(\lambda\) coefficient of 0.003. During inference-time steered generation, we found that setting \(\eta = 0.8\) produce suitable responses. 
\eat{
We study two experimental setups for steering towards a desired style: \textit{(i)} steering from a source style and \textit{(ii)} steering from LLM output that may belong to any kind of style.}

\subsection{Baselines}
We evaluate our approach against two main baselines: few-shot in-context learning and direct query-steering.

For few-shot in-context learning (ICL), we include N examples sampled from the training data in our prompts and apply a few-shot learning approach~\cite{brown2020language} to assess how well the pre-trained base model captures the intended transformation. Due to constraints in context length, we prefer to use \emph{phi-3-mini-128K instruct} model for this experiment.~\footnote{We have not found any claim that confirms \emph{phi-3-mini-128K instruct} has a better performance compared to its 4K counter part except increased input context window.}

To examine the impact of the sparse representation learned by SAEs, we apply our steering procedure directly to the query embedding. Specifically, we use the query embedding of the fine-tuned base model to guide the transformation without leveraging the sparse encoding.

\eat{
investigated independent query attention behavior to evaluate SAE's effectiveness in steering responses toward specific styles. Using the fine tuned base model's query embeddings for steering revealed good potential for directional control.
}

%% file: evaluate.tex
\begin{figure*}[ht]
\centering
  \includegraphics[width=.88\linewidth]{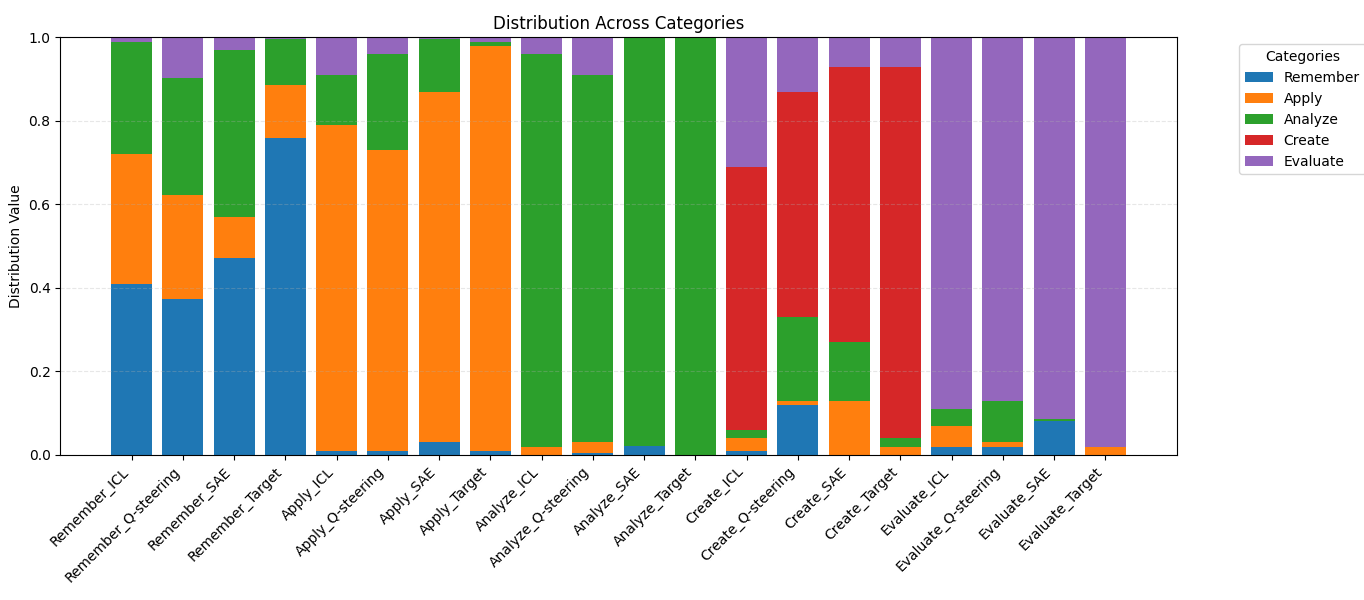}
  \caption{ Measured class distribution showing steering accuracy from source (Understand) to target cognitive categories. Target classes appear on X-axis while Y-axis shows class category distribution of steered text. }
  \label{fig:l1}
\end{figure*}

\section{Cognitive Transformation}

In this section, we evaluate our approach based on two evaluation methods: 1) Classification accuracy. 2) Categorizing steering-modified generated text. \eat{For a full list of experiments please also see \ref{sec: be}.}

\subsection{Quality of Sparse Encodings}
First, we evaluated the quality of the sparse embedding using a classification task. We created 
a few-shot learning framework with $N=30$ support examples per class to examine how well $z$ maintains distinction between different cognitive levels. Table \ref{tab:matching-analysis} suggests $L_{1}$ regularization and layers influence class separation. Increasing the $L_{1}$ coefficient generally led to more pronounced sparsity, improving differentiation between classes. Among the layers, middle layer displays the clearest benefit, indicating that sparsity at this intermediate representation level best captures distinct cognitive styles.

\begin{table}[htbp]
    \centering
    \caption{Layer-wise Analysis of Multi-class Classification}
    \label{tab:matching-analysis}
    \setlength{\tabcolsep}{4.2pt}
    \tiny
    \begin{tabular}{@{}lcc|ccccccc@{}}
        \toprule
        Layer & Coef. & N & rem & und & app & ana & eva & cre & avg \\
        \midrule
        \multirow{3}{*}{5} 
            & .0003 & 30 & \calcpercent{450}{960} & \calcpercent{313}{960} & \calcpercent{142}{960} & \calcpercent{51}{960} & \calcpercent{103}{960} & \calcpercent{216}{960} & 22.14\% \\
            & .0003 & 300 & \calcpercent{287}{960} & \calcpercent{391}{960} & \calcpercent{185}{960} & \calcpercent{154}{960} & \calcpercent{412}{960} & \calcpercent{292}{960} & 29.88\% \\
            & .003 & 30 & \calcpercent{497}{960} & \calcpercent{281}{960} & \calcpercent{137}{960} & \calcpercent{57}{960} & \calcpercent{90}{960} & \calcpercent{127}{960} & 20.64\% \\
            & .003 & 300 & \calcpercent{293}{960} & \calcpercent{344}{960} & \calcpercent{142}{960} & \calcpercent{130}{960} & \calcpercent{265}{960} & \calcpercent{65}{960} & 21.51\% \\
            & .03 & 30 & \calcpercent{464}{960} & \calcpercent{259}{960} & \calcpercent{126}{960} & \calcpercent{42}{960} & \calcpercent{84}{960} & \calcpercent{98}{960} & 18.63\% \\
            & .03 & 300 & \calcpercent{293}{960} & \calcpercent{344}{960} & \calcpercent{142}{960} & \calcpercent{130}{960} & \calcpercent{265}{960} & \calcpercent{65}{960} & 21.51\% \\
        \midrule
        \multirow{4}{*}{15} 
            & .0003 & 30 & \calcpercent{146}{960} & \calcpercent{746}{960} & \calcpercent{117}{960} & \calcpercent{139}{960} & \calcpercent{219}{960} & \calcpercent{427}{960} & 31.15\% \\
            & .0003 & 300 & \calcpercent{500}{960} & \calcpercent{344}{960} & \calcpercent{383}{960} & \calcpercent{237}{960} & \calcpercent{512}{960} & \calcpercent{507}{960} & 43.11\% \\
            & .003 & 30 & \calcpercent{164}{960} & \calcpercent{604}{960} & \calcpercent{101}{960} & \calcpercent{180}{960} & \calcpercent{232}{960} & \calcpercent{338}{960} & 28.11\% \\
            & .003 & 300 & \calcpercent{373}{960} & \calcpercent{288}{960} & \calcpercent{373}{960} & \calcpercent{225}{960} & \calcpercent{442}{960} & \calcpercent{472}{960} & 37.73\% \\
            & .03 & 30 & \calcpercent{144}{960} & \calcpercent{514}{960} & \calcpercent{119}{960} & \calcpercent{146}{960} & \calcpercent{270}{960} & \calcpercent{468}{960} & 28.84\% \\
            & .03 & 300 & \calcpercent{342}{960} & \calcpercent{274}{960} & \calcpercent{357}{960} & \calcpercent{228}{960} & \calcpercent{442}{960} & \calcpercent{501}{960} & 37.22\% \\
        \midrule
        \multirow{3}{*}{28} 
            & .0003 & 30 & \calcpercent{496}{960} & \calcpercent{407}{960} & \calcpercent{113}{960} & \calcpercent{61}{960} & \calcpercent{109}{960} & \calcpercent{189}{960} & 23.87\% \\
            & .0003 & 300 & \calcpercent{295}{960} & \calcpercent{282}{960} & \calcpercent{246}{960} & \calcpercent{181}{960} & \calcpercent{466}{960} & \calcpercent{308}{960} & 30.87\% \\
            & .003 & 30 & \calcpercent{347}{960} & \calcpercent{480}{960} & \calcpercent{105}{960} & \calcpercent{94}{960} & \calcpercent{100}{960} & \calcpercent{168}{960} & 22.47\% \\
            & .003 & 300 & \calcpercent{243}{960} & \calcpercent{259}{960} & \calcpercent{249}{960} & \calcpercent{164}{960} & \calcpercent{331}{960} & \calcpercent{97}{960} & 23.30\% \\
            & .03 & 30 & \calcpercent{282}{960} & \calcpercent{414}{960} & \calcpercent{85}{960} & \calcpercent{111}{960} & \calcpercent{76}{960} & \calcpercent{102}{960} & 18.58\% \\
            & .03 & 300 & \calcpercent{198}{960} & \calcpercent{219}{960} & \calcpercent{239}{960} & \calcpercent{144}{960} & \calcpercent{311}{960} & \calcpercent{82}{960} & 20.71\% \\
        \bottomrule
    \end{tabular}
    \\ \vspace{1mm}
    \footnotesize{Note: N=sample size; rem=remember; und=understand; app=apply; ana=analyze; eva=evaluate; cre=create; avg=average}
\end{table}
Additionally, augmenting the number of samples (N) used to compute class prototypes enhances the classification accuracy, suggesting that richer exemplars help the model refine its notion of each cognitive category. 

\subsection{Effectiveness of Transformation}

This evaluation involves classifying the steered text into Bloom's Taxonomy levels. Figure \ref{fig:l1} shows a bar chart comparing text generated using three steering approaches—baseline steering, Q-Steering, and SAE-based steering—as the style shifts from source class('Understand') to other styles, measured against the target data for each intended styles. Each bar in the chart represents the distribution of categories identified by Claude from the steered generations.

For example, we observe that query attention features are effective in steering text. However, due to the dense nature of query features, the generated sentences often exhibit multiple styles simultaneously. For instance, feedback from query embedding steering may include characteristics of both evaluate and apply levels of Bloom's Taxonomy. Enforcing sparsity in the feature space allows the model to better navigate toward specific attributes, as distinguishable attributes become more accessible in a sparse feature space.

Please refer to section~\ref{sec:Qualitative} for qualitative examples on SAE generated steered text, also refer to Section~\ref{sec:Qgeneration} for a qualitative comparison between with and without SAE  generated steered text.

\subsection{Different Layers of Attention}

\begin{figure}[!ht]
\centering
  \includegraphics[width=.88\linewidth]{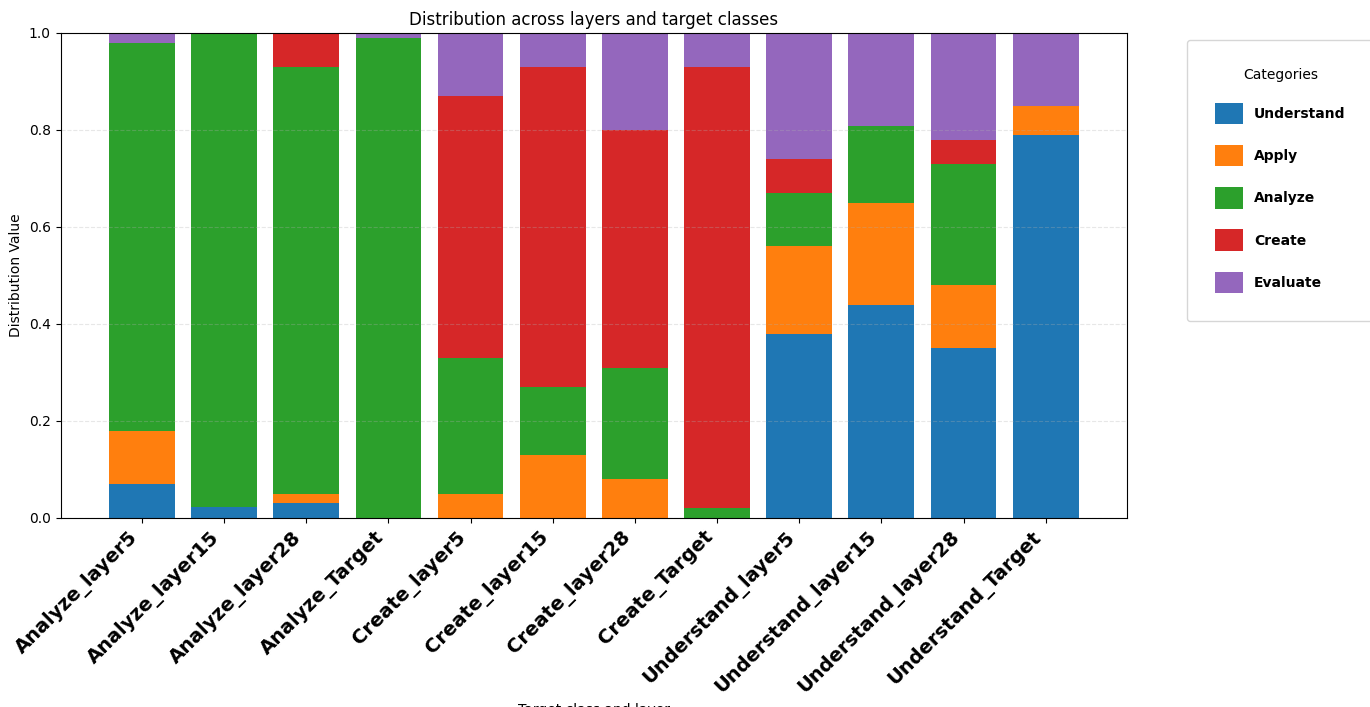}
  \caption{ Measured class distribution showing steering accuracy from source (Remember) to target cognitive categories. Target classes appear on X-axis while Y-axis shows class category distribution of steered text. }
  \label{fig:across_attn}
\end{figure}

We investigate the relationship between query embedding sparsity and steering behavior across different layers. In general, we find that sparse query features are better than any other layer components for steering. Middle layer attention demonstrates superior performance for steering tasks relative to other layers, as shown in the 'remember' to 'understand' task (last 4 bar charts) in Figure~\ref{fig:across_attn}. In contrast, features from lower layer attention (simpler and more localized relationships) and higher layer attention (specific and distant relationships) capture less of the intended target style attribute. This finding aligns with previous research showing that attention most strongly correlates with syntactic dependency relations in the model's middle layers~\cite{vig2019analyzing}, where these dependencies have their most effective representation. Further support comes from early studies~\cite{hu2021syntax,gong2020rich} demonstrating that syntactic dependency enhances textual style transfer by preserving content through grammatical structures.

\subsection{Impact of Sparsity Regularization}
\begin{figure}[!ht]
\centering
\includegraphics[width=.88\linewidth]{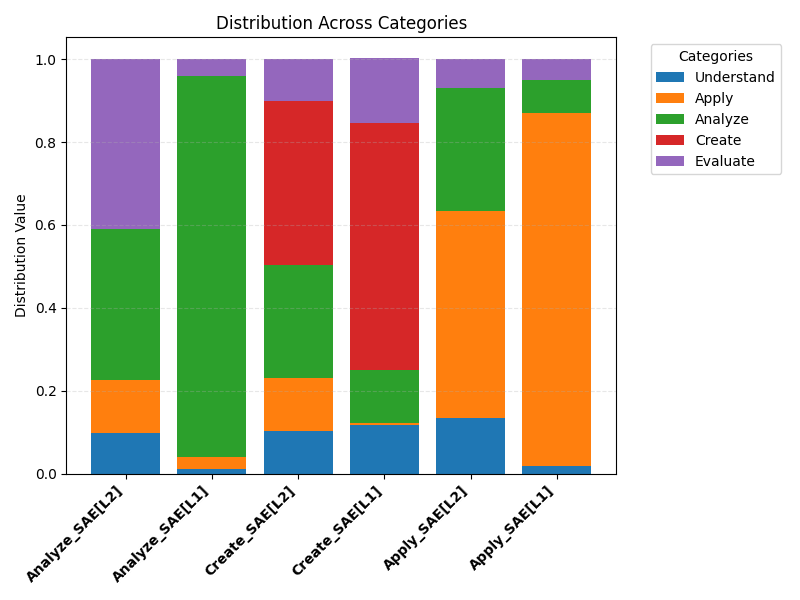}
\caption{Distribution of steering accuracy showing how effectively samples from the source class (Remember) were redirected to each intended target class, comparing SAE models trained with $L_{2}$ versus $L_{1}$ loss}
\label{fig:l2vl1}
\end{figure}

Training SAEs with $L_{2}$ versus $L_{1}$ regularization techniques creates fundamentally different feature spaces. $L_{2}$ regularization, which does not reduce features to zero, creates a less sparse feature space that results generated steered text, blending categories from different styles. This is evident in Figure~\ref{fig:l2vl1}'s 'remember' to 'analyze' task (first two bar charts), where only few samples are classified as analyze. In contrast, $L_{1}$ regularization creates sparse feature space that directs generation more strongly toward the intended target style, as sparsity preserves the distinct attribute while keeping unrelated features suppressed.

We explored using alternative model components beyond query attention heads for generation control. Specifically, we trained SAE on the residual stream at layers 5,15,25 but observed no meaningful impact on steering the output. This aligns with our intuition, as residual streams aggregate information from previous layers, potentially diluting the precise control needed for targeted content modulation. In contrast to middle layer, We observe features at layer 25 generates meaningful text. The qualitative comparison (remember to create) between SAE trained on query activations (SAE-Q) and residual activations (SAE-R) is shown in Figure~\ref{fig:responses}. The generated output from the SAE-R is inadequate, as it fails to properly align with the desired style.
\begin{figure}[!htbp]
{\small
\hrule
\vspace{.1cm}
Original Response (Remember): Your code shows awareness of function declarations and reference parameters, but there's a syntax error in the parameter declaration. The correct syntax for reference parameters in C++ is 'int\&' not '\&int'. This fundamental syntax is crucial for the code to compile properly.
\vspace{.1cm}
\hrule
\vspace{.1cm}
SAE-R: Your code demonstr understandingareness of pointer structure and pointer parameters, but there's a fundamental error in the pointer pattern. The relationship relationship should passing parameter requires C++ is '\&*', instead 'int'
\vspace{.1cm}
\hrule
\vspace{.1cm}
SAE-Q: While the function structure using reference parameter shows good understanding of memory concepts, the implementation fails in practical application. The relationship between function parameters and their effects on caller scope needs consideration.
\vspace{.1cm}
\hrule
\vspace{.1cm}
}
\caption{Comparison of responses between SAE trained on residual stream and SAEs trained on query attention feature.}
\label{fig:responses}
\end{figure}

\eat{
\medskip

\begin{lstlisting}
Original response (Remember):Your code 
shows awareness of function declarations 
and reference parameters, but there's a 
syntax error in the parameter 
declaration. The correct syntax for 
reference parameters in C++ is 'int\&' 
not '\&int'. This fundamental syntax is 
crucial for the code to compile properly.
\end{lstlisting}

\begin{lstlisting} 
SAE-R: Your code demonstr 
understandingareness of pointer 
structure and pointer parameters, but 
there's a fundamental error in the 
pointer pattern. The relationship 
relationship should passing parameter 
requires C++ is '\&*', instead ' 'int'

\end{lstlisting}

\begin{lstlisting}
SAE-Q: While the function structure
using reference parameter shows good 
understanding of memory concepts, the 
implementation fails in practical 
application. The relationship between 
function parameters and their effects on 
caller scope needs consideration.
\end{lstlisting}
}




%% file: bg.tex
\subsection{Sparse Autoencoders}
\label{sec:sae_bg}
The concept of sparse coding using over-complete dictionaries was first introduced by Mallat et al.~\cite{mallat1993matching}. This foundation was significantly advanced by Olshausen et al.~\cite{olshausen1996emergence}, introduced an unsupervised approach to learn the dictionary directly from data. While this method primarily used in image processing ~\cite{liu2015sparse}, a parallel development emerged when~\citet{hinton2006reducing} introduced the autoenoder architecture for dimensionality reduction. These two streams of research eventually converged in the development of SAE ~\cite{konda2014zero,lee2007sparse}, which incorporates sparsity constraints, such as $L_{1}$ regularization, to extract meaningful sparse features.

SAEs have recently gained popularity due to their ability to decompose LLM activations into interpretable features~\cite{cunningham2023sparse}. This decomposition enables the representation of semantic and syntactic concepts in LLMs as linear combinations of activation patterns, a property that has been effectively demonstrated across various language models, including BERT~\cite{yun2021transformer}, GPT~\cite{sharkey2022taking}, and Pythia~\cite{cunningham2023sparse}~\cite{bloom2024open,marks2023some}. The interpretability of sparse activations is grounded in the Linear Representation Hypothesis, which posits that these activations can be understood as linear combinations of features~\cite{mikolov2013linguistic,nanda2023progress,park2023linear}. For optimal interpretability, it is desirable for each feature to correspond to a single neuron, a concept known as mono-semantic neurons~\cite{bau2020understanding,goh2021multimodal}. However, this ideal scenario is often unattainable because neural networks must encode more features than there are available neurons. This limitation leads to the emergence of poly-semantic neurons, where a single neuron gets triggered by multiple, potentially unrelated features~\cite{scherlis2022polysemanticity,mu2020compositional}. This phenomenon is explained by the superposition hypothesis, which suggests that neurons share representational capacity by encoding multiple features in overlapping ways~\cite{elhage2022toy}. This trade-off between interpretability and representational efficiency underscores the challenges and opportunities in understanding LLM activations through SAEs. Further, methods like activation pruning techniques~\cite{rajamanoharan2024jumping} or thresholds~\cite{rajamanoharan2024improving} for feature selection, have helped improve the balance between keeping representations sparse while maintaining good reconstruction quality. 

\par
\subsection{Steered generation}

Research in computer vision and natural language processing (NLP) has shown that content generation can be controlled through adjustments in latent space~\cite{larsen2016autoencoding,white2016sampling} and word embeddings~\cite{han2024word}. Instead many popular approaches (\ie supervised fine-tuning, reinforcement learning from human feedback (RLHF), steerable layers, weight editing~\cite{ranzato2015sequence,dathathri2019plug,ilharco2022editing,meng2022locating}) involve modifying the model's weights directly, undermining the model's overall performance~\cite{qi2023fine}. Alternative methods that avoid direct weight modification and use intervention at different levels  like decoding~\cite{gu2017trainable}, prompt~\cite{zhou2022steering}, and token embeddings~\cite{khashabi2021prompt} have gained popularity for influencing model behavior without compromising performance.

\par
Intervening into a language model’s activations has emerged as an efficient strategy for controlling model behavior without modifying its weights. This approach involves learning steering vectors which guide the model's behavior more effectively than direct weight alterations~\cite{hernandez2023inspecting,subramani2022extracting}. Building on this idea, methods such as activation engineering, which leverage latent space arithmetic, have gained traction as practical tools for manipulating LLMs~\cite{correia2019adaptively}. However, the effectiveness of these approaches still relies on determining the steering vectors. Inference-Time Intervention (ITI)~\cite{li2024inference} computes steering vectors applied uniformly across all sequence positions during inference, using trained linear probes to identify attention heads with distinct activation patterns for true and false statements, enabling the model to produce more truthful outputs. ActAdd~\cite{turner2023activation} creates steering vectors by computing the difference in model activations between contrasting prompts, then adds these vectors during inference to guide the model's outputs toward desired traits. Similar approaches can also improve upon in-context learning by introducing in-context vectors~\cite{liu2023context}. While training SAE is computationally intensive~\cite{gao2024scaling}, their highly interpretable continuous domain makes them valuable for feature-based steering~\cite{durmusevaluating}. For instance, sparse activation editing in the middle layer's residual stream during inference time allows pre-trained SAE to effectively steer LLMs to resolve context-memory knowledge conflicts~\cite{zhao2024steering}.  By enforcing sparsity constraints, SAE facilitates the identification of distinct and meaningful features within the data~\cite{chalnev2024improving,o2024steering}.

We explore the role of sparsity in feature space  for steering text toward desired styles through independently trained SAEs on query attention heads. Unlike previous approaches using activation editing, we employ inference time gradient descent to achieve style transfer. Our findings align with layer-based feature patterns from previous works, particularly revealing that middle layer features enable SAEs to learn task-specific useful characteristics.

\eat{
Based on  the potential of LLM activations, various interpretability studies~\cite{burns2022discovering,nanda2023progress} include these activations or train models like SAE using these activations as input. Intuitively, 
These are generally considered more interpretable~\cite{marks2023interpreting} than LLM weights as they often correspond more directly to semantic concepts and features of the input~\cite{zou2023representation}. Recent findings also suggests that~\cite{zhao2024steering} signals of knowledge conflict can be detected in the residual stream of LLMs, particularly in the mid-layers. Our findings partially align with previous research showing that middle layers play an important role in steering text generation. However, unlike prior work, we found that attention layers - specifically query attention heads - were more influential than residual layers for understanding text attributes~\cite{vig2019analyzing} . Our approach focuses on creating sparse attention activations, which effectively guide the model toward generating text with the desired style.

\par

}

%% file: appendix.tex
\section{Experiments on Mistral}
\label{sec:mistral}
To verify the consistency of our approach, we conduct experiments using another model—Mistral 7b instruct v0.1 (~\cite{jiang2023mistral}). The results  presented below indicate that our approach remains consistent across different models.

\begin{table}[htbp]
    \centering
    \caption{Layer-wise Analysis of Multi-class Classification}
    \label{tab:matching-analysis_appendox}
    \setlength{\tabcolsep}{6.8pt}
    \scriptsize
    \begin{tabular}{@{}c@{\,}c@{\,}|@{\,}c@{\,}c@{\,}c@{\,}c@{\,}c@{\,}c@{\,}c@{}}
        \toprule
        Layer & N & rem & und & app & ana & eva & cre & avg \\
        \midrule
        15 & 30 & \calcpercent{181}{960} & \calcpercent{361}{960} & \calcpercent{86}{960} & \calcpercent{126}{960} & \calcpercent{60}{960} & \calcpercent{760}{960} & 27.3\% \\
        & 300 & \calcpercent{319}{960} & \calcpercent{327}{960} & \calcpercent{178}{960} & \calcpercent{159}{960} & \calcpercent{536}{960} & \calcpercent{694}{960} & 38.4\% \\
        & 950 & \calcpercent{303}{960} & \calcpercent{431}{960} & \calcpercent{299}{960} & \calcpercent{15}{960} & \calcpercent{344}{960} & \calcpercent{593}{960} & 34.5\% \\
        \bottomrule
    \end{tabular}
    \\ \vspace{1mm}
    \footnotesize{Note: N=sample size; rem=remember; und=understand;
    app=apply; ana=analyze; eva=evaluate; cre=create; avg=average}
\end{table}
\begin{figure*}[ht]
\centering
  \includegraphics[width=.88\linewidth]{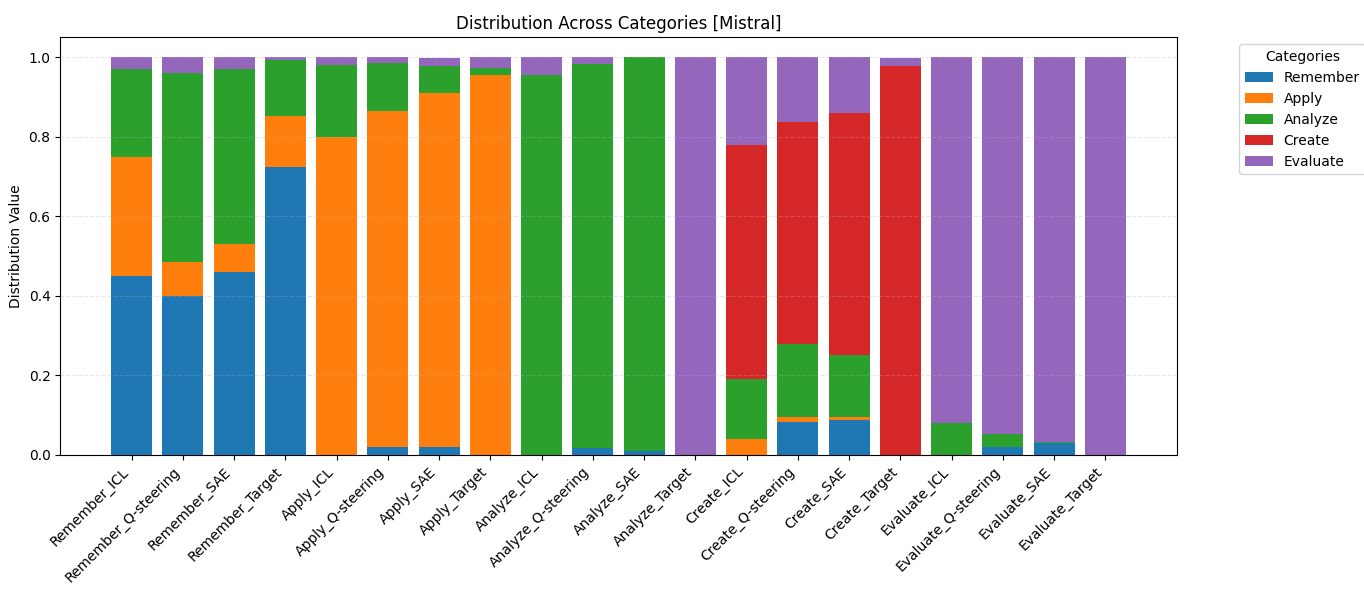}
  \caption{Measured class distribution showing steering accuracy from source (Understand) to target cognitive categories for \emph{Mistral 7b Instruct} model. Target classes appear on X-axis while Y-axis shows class category distribution of steered text.}
  \label{fig:mistral}
\end{figure*}

Figure \ref{fig:mistral} shows the distribution of steered generation category using mistral.

\subsection{Qualitative comparison using Mistral}
\medskip
\subsection*{Understand to apply}
\textbf{Query:}

Below is an instruction that describes a task, paired with an input that provides further context. Write a response that appropriately completes the request.\par
\textbf{Instruction:}\par
Provide feedback on the following code \par
Write a function that converts alternate characters to uppercase in a string using call by reference.\par

\textbf{Code}:
\begin{lstlisting}
#include <iostream>
#include <string>
using namespace std;
void alternateCase(string &str) {
    for(int i = 1; i < str.length(); i += 2) {
        str[i] = toupper(str[i]);
    }
}
int main() {
    string text = "programming";
    cout << "Original: " << text << endl;
    alternateCase(text);
    cout << "Modified: " << text;
    return 0;
}
\end{lstlisting}
\textbf{W/O SAE (Q generation):} The relationship between the string length and the loop boundary shows basic grasp of character access and array bounds. The implementation demonstrates fundamental concepts of C++ syntax while maintaining proper pointer usage for in-place string modification. \par

\textbf{SAE:} While the approach is efficient for small-to-medium string sizes, the single-character modification strategy limits scalability. This implementation might not perform well for larger strings or handle multiple consecutive consecutive characters.  \par
\textbf{Judgement:}  SAE Generation is more effective in aligning with the "Evaluate" level of Bloom's Taxonomy than Q Generation . The "Evaluate" level involves making judgments or decisions based on specific criteria or standards.

In SAE Generation, the feedback provides a critical evaluation of the approach by identifying its potential limitations. It assesses the performance and scalability concerns for larger strings or scenarios with multiple consecutive characters. This evaluation is based on the specific criteria of efficiency, scalability, and edge-case handling, which are essential characteristics of a robust implementation.

\medskip
\subsection*{Understand to create}
\textbf{Query:}

Below is an instruction that describes a task, paired with an input that provides further context. Write a response that appropriately completes the request.\par
\textbf{Instruction:}\par
Provide feedback on the following code \par
Write a program to swap two numbers using call by reference method.\par

\textbf{Code}:
\begin{lstlisting}
#include <iostream>
using namespace std;
void swapNumbers(int &x, int &y) {
    x = y;
    y = x;
}
int main() {
    int a = 10, b = 20;
    cout << "Before swap: a = " << a << ", b = " << b << endl;
    swapNumbers(a, b);
    cout << "After swap: a = " << a << ", b = " << b << endl;
    return 0;
}
\end{lstlisting}

\textbf{W/O SAE (Q generation):} The implementation shows basic understanding of how references work in C++, but the fundamental relationship between value and reference parameters needs clearer expression. A proper function signature (int$\&$) for a reference parameter should follow the syntax: *reference(value). \par

\textbf{SAE:}  Consider enhancing the function by adding input validation for zero or negative numbers, and implementing a template version for generic number types. You might also look into implementing swapping operations in different programming languages to compare their syntax and performance. \par

\textbf{Judgement:}  SAE Generation  aligns more effectively with the "create" level of Bloom's Taxonomy than Generation 1. The "create" level involves putting parts together to form something new or original, and Generation 2 provides suggestions for extending and enhancing the existing implementation in novel ways.

In contrast, Q Generation  primarily focuses on providing feedback on the existing implementation and suggesting improvements to the function signature syntax. While this feedback is valuable, it does not explicitly encourage creating new elements or substantially modifying the existing implementation.

\medskip
\subsection*{Understand to analyze}
\textbf{Query:}

Below is an instruction that describes a task, paired with an input that provides further context. Write a response that appropriately completes the request.\par
\textbf{Instruction:}\par
Provide feedback on the following code \par
Write a function that removes duplicate elements from a sorted vector while maintaining the original order.\par

\textbf{Code}:
\begin{lstlisting}
#include <iostream>
#include <vector>
using namespace std;

vector<int> removeDuplicates(vector<int>& nums) {
    vector<int> result;
    for(int i = 1; i < nums.size(); i++) {  
        if(nums[i] != nums[i-1]) {
            result.push_back(nums[i]);
        }
    }
    return result;
}
\end{lstlisting}
\textbf{W/O SAE (Q generation):}  Your code effectively utilizes the vector's constructor for memory allocation and push\_back for appending, demonstrating fundamental vector manipulation using C++ syntax. The logic correctly maintains the order but loses duplicates, showing how small details can have significant differences.\par

\textbf{SAE:}  Consider implementing a simple hash table for O(N) time complexity using a hash table's constant access time for duplicate detection. This would provide better performance and a more elegant solution for the same functionality, but at the cost of linear probing for collisions. \par
\textbf{Judgement:} SAE Generation  is more effective than Generation 1 in terms of aligning with the "Analyze" level of Bloom's Taxonomy because it requires the learner to break down the problem into its constituent parts, evaluate the pros and cons of different approaches, and make connections between the solution and its performance characteristics.

In contrast, Q Generation  primarily focuses on describing the existing code and its functionality, which aligns more with the "Understand" level of Bloom's Taxonomy. While it mentions the impact of losing duplicates, it does not explicitly prompt the learner to analyze the problem from different perspectives or evaluate alternative solutions.

\medskip
\subsection*{Understand to evaluate}

\textbf{Query:}

Below is an instruction that describes a task, paired with an input that provides further context. Write a response that appropriately completes the request.\par
\textbf{Instruction:}\par
Provide feedback on the following code \par
Write a function that converts alternate characters to uppercase in a string using call by reference. \par

\textbf{Code}:
\begin{lstlisting}
#include <iostream>
#include <string>
using namespace std;
void alternateCase(string &str) {
    for(int i = 1; i < str.length(); i += 2) {
        str[i] = toupper(str[i]);
    }
}
int main() {
    string text = "programming";
    cout << "Original: " << text << endl;
    alternateCase(text);
    cout << "Modified: " << text;
    return 0;
}
\end{lstlisting}
\textbf{W/O SAE (Q generation):} The relationship between the string length and the loop boundary shows basic grasp of character access and array bounds. The implementation demonstrates fundamental concepts of C++ syntax while maintaining proper pointer usage for in-place string modification. \par

\textbf{SAE:}  While the approach is efficient for small-to-medium string sizes, the single-character modification strategy limits scalability. This implementation might not perform well for larger strings or handle multiple consecutive consecutive characters.  \par
\textbf{Judgement:} 

In SAE Generation , the feedback provides a critical evaluation of the approach by identifying its potential limitations. It assesses the performance and scalability concerns for larger strings or scenarios with multiple consecutive characters. This evaluation is based on the specific criteria of efficiency, scalability, and edge-case handling, which are essential characteristics of a robust implementation.

On the other hand, Q Generation  primarily focuses on describing the implementation's functionality and adherence to syntax rules. While this feedback demonstrates an understanding of the code, it does not provide a critical evaluation or make judgments based on specific criteria, which is a key aspect of the "Evaluate" level.

\section{Experiments on Query embeddings}
\label{sec:Q}
We also investigate how different cognitive styles ( Table \ref{tab:matching-analysis_appendox}) are distinct in the query embedding space across both the models mistral and phi. 
\begin{table}[htbp]
\begin{minipage}{\columnwidth}
    \centering
    \caption{Layer-wise Analysis of Multi-class Classification w/o SAE with fixed $L_{1}$=0.003}
    \label{tab:matching-analysis_appendox}
    \setlength{\tabcolsep}{5.5pt}
    \scriptsize
    \begin{tabular}{@{}l@{\,}c@{\,}c@{\,}|@{\,}c@{\,}c@{\,}c@{\,}c@{\,}c@{\,}c@{\,}c@{}}
        \toprule
        L & N & mod & rem & und & app & ana & eva & cre & avg \\
        \midrule
        \multirow{4}{*}{15} 
            & 30 & mis & \calcpercent{266}{960} & \calcpercent{609}{960} & \calcpercent{77}{960} & \calcpercent{149}{960} & \calcpercent{218}{960} & \calcpercent{722}{960} & 35.4\% \\
            & 300 & mis & \calcpercent{492}{960} & \calcpercent{291}{960} & \calcpercent{351}{960} & \calcpercent{280}{960} & \calcpercent{595}{960} & \calcpercent{777}{960} & 48.4\% \\
            & 30 & phi & \calcpercent{288}{960} & \calcpercent{548}{960} & \calcpercent{196}{960} & \calcpercent{214}{960} & \calcpercent{443}{960} & \calcpercent{547}{960} & 38.8\% \\
            & 300 & phi & \calcpercent{537}{960} & \calcpercent{380}{960} & \calcpercent{452}{960} & \calcpercent{195}{960} & \calcpercent{507}{960} & \calcpercent{601}{960} & 46.4\% \\
        \midrule
        \multirow{2}{*}{28} 
            & 30 & phi & \calcpercent{242}{960} & \calcpercent{512}{960} & \calcpercent{87}{960} & \calcpercent{125}{960} & \calcpercent{116}{960} & \calcpercent{350}{960} & 24.9 \\
            & 300 & phi & \calcpercent{278}{960} & \calcpercent{281}{960} & \calcpercent{343}{960} & \calcpercent{153}{960} & \calcpercent{451}{960} & \calcpercent{315}{960} & 31.6 \\
        \bottomrule
    \end{tabular}
    \\ \vspace{1mm}
    \footnotesize{Note: L=layer; N=sample size; mod=model (mis=mistral, phi=phi);
    rem=remember; und=understand; app=apply; ana=analyze; eva=evaluate; cre=create}
    \end{minipage}
\end{table}

\section{Experiments on SAE with different dimensions}
\label{sec:Q_dim}

\begin{figure*}[ht]
\centering
  \includegraphics[width=.88\linewidth]{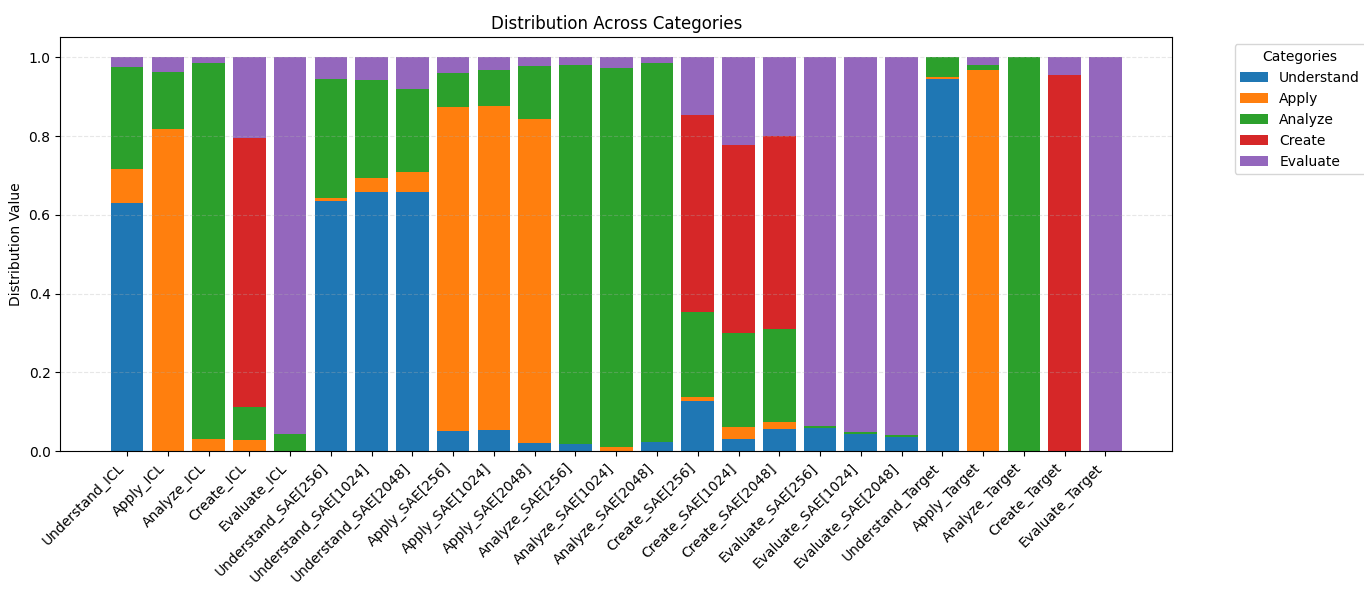}
  \caption{ Measured class distribution showing steering accuracy from source cognitive style (Remember) to intended target cognitive styles. Target classes appear on X-axis while Y-axis shows class category distribution of steered text.}
  \label{fig:acd}
\end{figure*}

Instead of only 512 dimension we also train SAE for different dimensions (\ie 256,1024,2048). Table \ref{tab:matching-analysis_appendox1} shows the classification accuracy across different dimensions. Figure \ref{fig:acd} shows the steering accuracy across different dimensions. We observe learning sparse features using the query attention's activations across various dimensions can be useful for textual attribute understanding. 

\begin{table}[htbp]
\begin{minipage}{\columnwidth}
    \centering
    \caption{Analysis of Multi-class Classification with SAE trained on different dimension with fixed $L_{1}$=0.003 and Layer 15}
    \label{tab:matching-analysis_appendox1}
    \setlength{\tabcolsep}{3.5pt}
    \scriptsize
    \begin{tabular}{@{}l@{\,}c@{\,}|@{\,}c@{\,}c@{\,}c@{\,}c@{\,}c@{\,}c@{\,}c@{}}
        \toprule
        dim & N & rem & und & app & ana & eva & cre & avg \\
        \midrule
         256 & 30  & \calcpercent{216}{960} & \calcpercent{640}{960} & \calcpercent{116}{960} & \calcpercent{200}{960} & \calcpercent{341}{960} & \calcpercent{502}{960} & 35.0\% \\
        
         256 & 300  & \calcpercent{506}{960} & \calcpercent{338}{960} & \calcpercent{416}{960} & \calcpercent{241}{960} & \calcpercent{508}{960}  & \calcpercent{558}{960} & 44.6\% \\
        
         1024 & 30  & \calcpercent{365}{960} & \calcpercent{287}{960} & \calcpercent{383}{960} & \calcpercent{227}{960} & \calcpercent{424}{960} & \calcpercent{456}{960} & 37.2\% \\
        
         1024 & 300  & \calcpercent{365}{960} & \calcpercent{287}{960} & \calcpercent{383}{960} & \calcpercent{227}{960} & \calcpercent{424}{960} & \calcpercent{456}{960} & 37.2\% \\
        
         2048 & 30  & \calcpercent{188}{960} & \calcpercent{576}{960} & \calcpercent{100}{960} & \calcpercent{181}{960} & \calcpercent{185}{960} & \calcpercent{273}{960} & 26.1\% \\
        
         2048 & 300  & \calcpercent{330}{960} & \calcpercent{282}{960} & \calcpercent{375}{960} & \calcpercent{219}{960} & \calcpercent{414}{960} & \calcpercent{435}{960} & 35.7\% \\
        \bottomrule
    \end{tabular}
    \\ \vspace{1mm}
    \footnotesize{Note: dim=dimension; N=sample size; rem=remember; und=understand;
    app=apply; ana=analyze; eva=evaluate; cre=create}
    \end{minipage}
\end{table}

\section{In context learning in baseline model}

\begin{figure}[ht]
\centering
  \includegraphics[width=.88\linewidth]{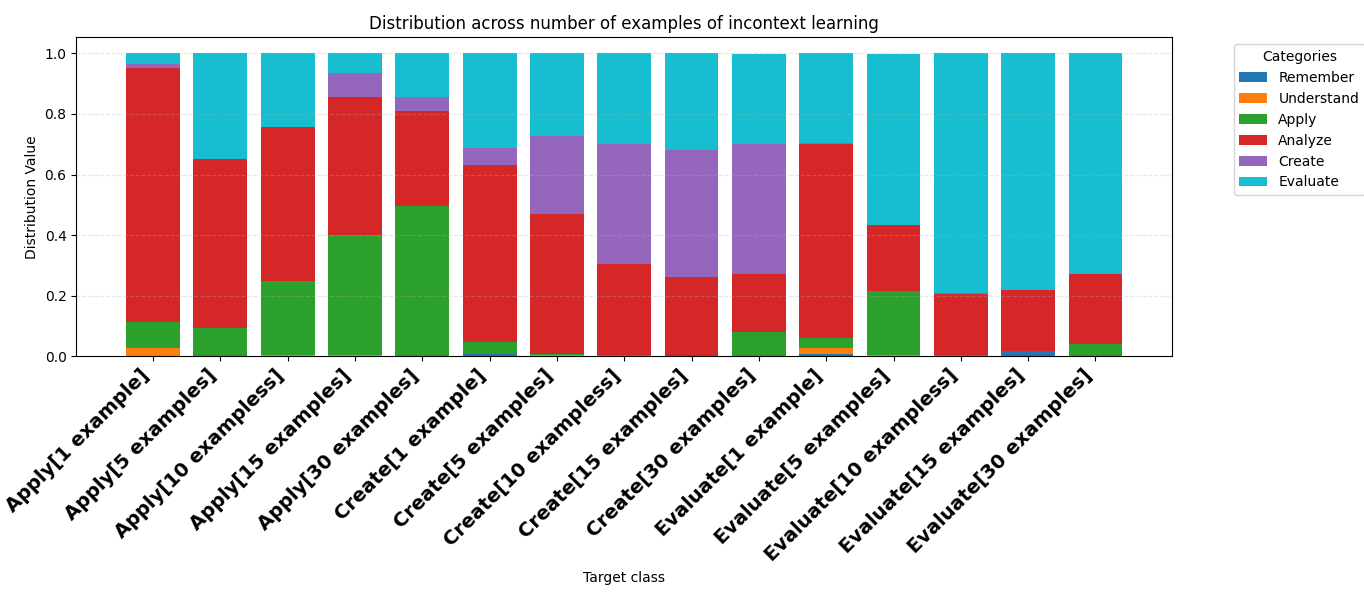}
  \caption{ Increasing the number of examples for a specific cognitive style improves the possibility of generating feedback in that style for base model.  }
  \label{fig:baselineincrement}
\end{figure}

Figure \ref{fig:baselineincrement} demonstrates how the baseline model's accuracy improves as the number of training examples increases. With only one example for the "create" cognitive style feedback, the model tends to generate responses that align more with "analyze" or "evaluate". However, as more examples are provided, the model becomes better at generating feedback in the intended "create" cognitive style.
Interestingly, the results also reveal that LLMs typically generate feedback in "evaluate" cognitive styles, even with just a single example of 'evaluate' feedback, the model generates majority of responses categories in this style. For this baseline implementation, we select the \emph{phi-3-mini-128k instruct model} primarily due to its extended context window, which allows us to process N examples simultaneously. While the literature contains no substantive claims comparing this model's performance to  \emph{phi-3-mini-4k instruct model}, except expanded context length capacity \cite{abdin2024phi}.

Below we present a qualitative study to demonstrate this effect. We are trying to steer the generation towards "create" cognitive level.

\section{Qualitative Examples}

\label{sec:Qualitative}
Below, we present examples of the generated responses after steering from source class (original response), accompanied by query and their corresponding target data.

\medskip
\subsubsection*{Steering from Remember to Create}
\textbf{Query:}

Below is an instruction that describes a task, paired with an input that provides further context. Write a response that appropriately completes the request.\par
\textbf{Instruction:}\par
Provide feedback on the following code\par
Write a function that calculates the sum of all elements in a vector of integers.\par
\textbf{Input}:\par
Code:
\begin{lstlisting}
#include <iostream>
#include <vector>
using namespace std;
int sumVector(const vector<int>& vec) {
    int sum = 0;
    for(int num : vec) {
        sum += num;
    }
    return sum;
}
int main() {
    vector<int> numbers = {1, 2, 3, 4, 5};
    cout << sumVector(numbers);
    return 0;
}
\end{lstlisting}

\textbf{Original response (Remember):} Your code demonstrates solid grasp of fundamental C++ elements including vector containers, range-based for loops, and function parameters. The basic syntax and structure show clear recognition of core programming constructs.

\textbf{Steered Generation (Create):} Consider enhancing the function with input validation, template support for different numeric types, and parallel processing capabilities for large vectors. Adding error handling and making it more generic could enhance its re usability.

\medskip
\subsubsection*{Steering from Remember to Analyze}
\textbf{Query:}\par

Below is an instruction that describes a task, paired with an input that provides further context. Write a response that appropriately completes the request.\par

\textbf{Instruction:}\par
Provide feedback on the following code\par
Write a function that calculates and returns the sum of all elements in a vector of integers.\par

\textbf{Code:}
\begin{lstlisting}
#include <iostream>
#include <vector>
using namespace std;

int vectorSum(vector<int>& vec) {
    int sum = 0;
    for(int num : vec) {
        sum += num;
    }
    return sum;
}

int main() {
    vector<int> numbers = {1, 2, 3, 4, 5};
    cout << vectorSum(numbers);
    return 0;
}
\end{lstlisting}

{\textbf{Original response (Remember):}  Your code correctly demonstrates fundamental vector operations using essential C++ syntax. The use of a range-based for loop shows basic iteration knowledge, while the parameter passing by reference indicates memory management awareness.}

\textbf{Steered Generation (Analyze):} This solution achieves both memory efficiency and computational simplicity with O(n) complexity. However, it lacks input validation and error handling, which could lead to issues with edge cases or special values. The choice of pass-by-reference over pass-by-value affects memory usage and performance.

\medskip

\subsubsection*{Steering from Remember to Evaluate}

\textbf{Query:}\par
Below is an instruction that describes a task, paired with an input that provides further context. Write a response that appropriately completes the request.
\par

\textbf{Instruction:}\par
Provide feedback on the following code\par
 Create a function that takes a vector of integers and returns the sum of all even numbers in the vector.
\par

\textbf{Code:}
\begin{lstlisting}
#include <iostream>
#include <vector>
using namespace std;

int sumEvenNumbers(vector<int> nums) {
    int sum;
    for(int num : nums) {    
        if(num % 2 == 0) {
            sum += num;
        }
    }
    return sum;
}

\end{lstlisting}

{\textbf{Original response (Remember):}  The code shows fundamental use of vector iteration and modulo operator, but overlooks a critical initialization of the sum variable which could lead to undefined behavior. The basic structure of using a for-each loop to traverse the vector is present.}

\textbf{Steered Generation (Evaluate):} While the code structure is clear and the solution is concise, the uninitialized sum variable introduces potential undefined behavior which could be problematic in production environments. The choice of using a range-based for loop shows good modern C++ practice.

\subsubsection*{Steering from Remember to Apply}

\textbf{Query:}\par
Below is an instruction that describes a task, paired with an input that provides further context. Write a response that appropriately completes the request.
\par

\textbf{Instruction:}\par
Provide feedback on the following code\par
Create a function that reverses the elements of a vector without using built-in reverse function.
\par

\textbf{Code:}
\begin{lstlisting}
#include <iostream>
#include <vector>
using namespace std;

void reverseVector(vector<int>& vec) {
    int left = 0, right = vec.size() - 1;
    while(left < right) {
        swap(vec[left], vec[right]);
        left++;
        right--;
    }
}

int main() {
    vector<int> numbers = {1, 2, 3, 4, 5};
    reverseVector(numbers);
    for(int num : numbers) {
        cout << num << " ";
    }
    return 0;
}
\end{lstlisting}

{\textbf{Original response (Remember):} Your code correctly implements the fundamental two-pointer technique, demonstrating solid understanding of vector indexing and basic swap operations. The core structure shows proper variable initialization and loop control flow essential for array manipulation.}

\textbf{Steered Generation (Apply):} The implementation effectively uses two pointers from both ends of the vector, performing swaps as the pointers converge. The relationship between the left and right pointers creates a symmetrical pattern, though the boundary condition could be more robust.

\subsubsection*{Steering from Remember to understand}

\textbf{Query:}\par
Below is an instruction that describes a task, paired with an input that provides further context. Write a response that appropriately completes the request.

\par

\textbf{Instruction:}\par
Provide feedback on the following code\par
Write a function that takes a vector of integers and returns the sum of all even numbers.

\textbf{Code:}
\begin{lstlisting}
#include <iostream>
#include <vector>
using namespace std;

int sumEvenNumbers(vector<int> nums) {  
    int sum = 0;
    for(int i = 1; i <= nums.size(); i++) {  
        if(nums[i] % 2 == 0) {
            sum += nums[i];
        }
    }
    return sum;
}
\end{lstlisting}

{\textbf{Original response (Remember):} The code shows fundamental array access and loop structures, but contains a critical indexing error. Array indices in C++ start at 0, not 1, and the loop condition should use < rather than <= to avoid buffer overflow.}

\textbf{Steered Generation (Understand):} The implementation attempts to sum even numbers but fails to properly apply vector indexing conventions. The relationship between vector size and valid indices is misunderstood, leading to incorrect access and potential buffer overflow issues.

\medskip
\section{Qualitative comparisons between with and without SAE based steering }

\label{sec:Qgeneration}
\subsection*{Understand to analyze}

\textbf{Query:}

Below is an instruction that describes a task, paired with an input that provides further context. Write a response that appropriately completes the request.\par
\textbf{Instruction:}\par
Provide feedback on the following code \par
Write a function that finds the length of the longest consecutive elements sequence in an unsorted vector.\par

\textbf{Code}:
\begin{lstlisting}
#include <iostream>
#include <vector>
#include <unordered_set>
using namespace std;
int longestConsecutive(vector<int>& nums) {
    unordered_set<int> num_set(nums.begin(), nums.end());
    int longest = 0;
    for(int num : nums) {
        if(num_set.find(num-1) == num_set.end()) {
            int currentNum = num;
            int currentStreak = 1;
            while(num_set.find(currentNum+1)) { 
                currentNum += 1;
                currentStreak += 1;
            }
            longest = max(longest, currentStreak);
        }
    }
    return longest;
}
\end{lstlisting}
\textbf{W/O SAE (Q generation):} Your code demonstrates solid grasp of fundamental data structures, particularly using unordered\_set for O(1) lookups. You've correctly identified basic elements like vector operations, loops, and conditional statements as building blocks for this problem.\par

\textbf{SAE:} While your solution works correctly, it uses more complex data structures than strictly necessary. The unordered\_set ensures uniqueness but might be more computationally expensive than a simple array or vector operation with appropriate tracking mechanisms.\par
\textbf{Judgement:} SAE Generation  is more effective than W/O SAE in terms of aligning with the "Analyze" level of Bloom's Taxonomy. The "Analyze" level involves breaking down information into parts, finding patterns and relationships, and identifying the underlying components or structure.

In SAE Generation, the feedback focuses on analyzing the solution in terms of its computational complexity and efficiency. It recognizes that while the solution works correctly, it employs more complex data structures (unordered\_set) than strictly necessary. This feedback demonstrates an understanding of the trade-offs involved in using different data structures and their impact on performance and resource utilization.

In contrast, W/O SAE generation ,primarily focuses on acknowledging the correctness of the solution and recognizing the use of basic programming constructs like data structures, loops, and conditionals. While this feedback is valuable in terms of understanding the solution's functionality, it lacks the depth of analysis required at the "Analyze" level of Bloom's Taxonomy.

\medskip
\subsection*{Understand to apply}

\textbf{Query:}

Below is an instruction that describes a task, paired with an input that provides further context. Write a response that appropriately completes the request.\par
\textbf{Instruction:}\par
Provide feedback on the following code \par
Create a function that rotates a vector to the right by k positions.\par

\textbf{Code}:
\begin{lstlisting}
#include <iostream>
#include <vector>
using namespace std;

void rotateVector(vector<int>& nums, int k) {
    int n = nums.size();
    k = k % n;
    vector<int> temp(n);
    for(int i = 0; i < k; i++) {
        temp[i] = nums[n-k+i];
    }
    for(int i = n-1; i >= k; i--) {
        nums[i] = nums[i-k];
    }
    for(int i = 0; i < k; i++) {
        nums[i] = temp[i];
    }
}

int main() {
    vector<int> nums = {1, 2, 3, 4, 5};
    rotateVector(nums, 2);
    for(int num : nums) {
        cout << num << " ";
    }
    return 0;
}

\end{lstlisting}
\textbf{W/O SAE (Q generation):} . Your code demonstrates solid grasp of fundamental concepts like vector manipulation and the modulo operator for handling rotation. The core structure using three distinct loops shows understanding of array traversal and value swapping mechanisms.\par

\textbf{SAE:} Consider implementing a left rotation by modifying the index calculation in the first loop to n-k. You might explore in-place rotation approaches using GCD or reversal algorithms to optimize the solution's computational efficiency. \par
\textbf{Judgement:} SAE Generation  is more effective than Q generation in aligning with the "Apply" level of Bloom's Taxonomy. The "Apply" level involves using learned material in new situations or applying concepts, rules, and principles to solve problems.

In Q generation, the feedback primarily focuses on evaluating the student's understanding and acknowledging their grasp of fundamental concepts. While this feedback is valuable, it does not explicitly challenge the student to apply their knowledge in a new or different way, which is the essence of the "Apply" level.

On the other hand, SAE Generation  provides specific suggestions and prompts for the student to consider alternative approaches and optimizations to the problem at hand. By suggesting the implementation of a left rotation using a modified index calculation or exploring in-place rotation approaches using algorithms like GCD or reversal, the feedback encourages the student to apply their existing knowledge and skills in a new context or problem scenario.
\medskip

\subsection*{Understand to create}

\textbf{Query:}

Below is an instruction that describes a task, paired with an input that provides further context. Write a response that appropriately completes the request.\par
\textbf{Instruction:}\par
Provide feedback on the following code \par
Write a function that reverses a string and changes lowercase to uppercase and vice versa using call by reference.\par

\textbf{Code}:
\begin{lstlisting}
#include <iostream>
#include <string>
using namespace std;
void reverseAndChangeCase(string str) {
    int n = str.length();
    for(int i = 0; i < n/2; i++) {
        char temp = str[i];
        str[i] = str[n-1-i];
        str[n-1-i] = temp;
    }
    for(int i = 0; i < n; i++) {
        if(isupper(str[i]))
            str[i] = tolower(str[i]);
        else
            str[i] = toupper(str[i]);
    }
}
int main() {
    string text = "Hello World";
    cout << "Original string: " << text << endl;
    reverseAndChangeCase(text);
    cout << "Modified string: " << text << endl;
    return 0;
}
\end{lstlisting}
\textbf{W/O SAE (Q generation):} The relationship between string length and index traversal shows careful consideration of symmetric elements, but the current implementation fails to maintain character consistency in case transformations. The logic behind converting characters needs refinement to avoid unnecessary operations.\par

\textbf{SAE:} The solution could be enhanced by combining the two loops into a single iteration, or by implementing a custom case conversion function to eliminate redundant checks. Adding string bounds checking or implementing a template-based approach might offer more flexibility. \par
\textbf{Judgement:} Q Generation primarily focuses on identifying issues and lacks specific suggestions for creating new structural or functional elements within the solution. While it identifies areas for improvement, it does not propose concrete ideas for reorganizing or combining existing elements into a new pattern or structure.

SAE Generation  better aligns with the "Create" level of Bloom's Taxonomy by providing specific examples and suggestions that involve reorganizing, combining, and introducing new elements to form a more coherent, efficient, and flexible solution.

\medskip
\subsection*{Understand to evaluate}

\textbf{Query:}

Below is an instruction that describes a task, paired with an input that provides further context. Write a response that appropriately completes the request.\par
\textbf{Instruction:}\par
Provide feedback on the following code \par
Write a function that reverses the elements of a vector.\par

\textbf{Code}:
\begin{lstlisting}
#include <iostream>
#include <vector>
using namespace std;

void reverseVector(vector<int>& nums) {
    int start = 0, end = nums.size();
    while(start < end) {
        swap(nums[start], nums[end]);
        start++;
        end--;
    }
}

int main() {
    vector<int> numbers = {1, 2, 3, 4, 5};
    reverseVector(numbers);
    for(int num : numbers) cout << num << " ";
    return 0;
}
\end{lstlisting}
\textbf{W/O SAE (Q generation):} .Your code demonstrates fundamental understanding of vector manipulation concepts, using pointer variables (start and end) to traverse the array from both ends. The basic swap operation and vector indexing show recognition of core memory concepts.\par

\textbf{SAE:} The code effectively implements the in-place reversal algorithm with appropriate boundary management. Your use of vector references and proper increment/decrement operations shows practical grasp of memory manipulation concepts. \par
\textbf{Judgement:} SAE Generation  is more effective than Q Generation in terms of aligning with the "Evaluate" level of Bloom's Taxonomy. The "Evaluate" level requires students to make judgments or decisions based on criteria and standards, which involves higher-order thinking skills such as critiquing, justifying, and assessing.

In SAE Generation, the feedback focuses on evaluating the practical implementation and effectiveness of the student's code. It specifically praises the "effective implementation of the in-place reversal algorithm with appropriate boundary management," which demonstrates the ability to assess the code's correctness and efficiency. Additionally, the feedback evaluates the student's "practical grasp of memory manipulation concepts" by highlighting their appropriate use of vector references and proper increment/decrement operations.

On the other hand, Q Generation  primarily focuses on describing the student's understanding and recognition of concepts, which aligns more with the "Understand" level of Bloom's Taxonomy. While it mentions the use of pointer variables and vector indexing, it does not provide an evaluation or critique of the code's implementation or effectiveness.

\section{Steering from fine tuned base model}

\begin{figure*}[ht]
\centering
  \includegraphics[width=.88\linewidth]{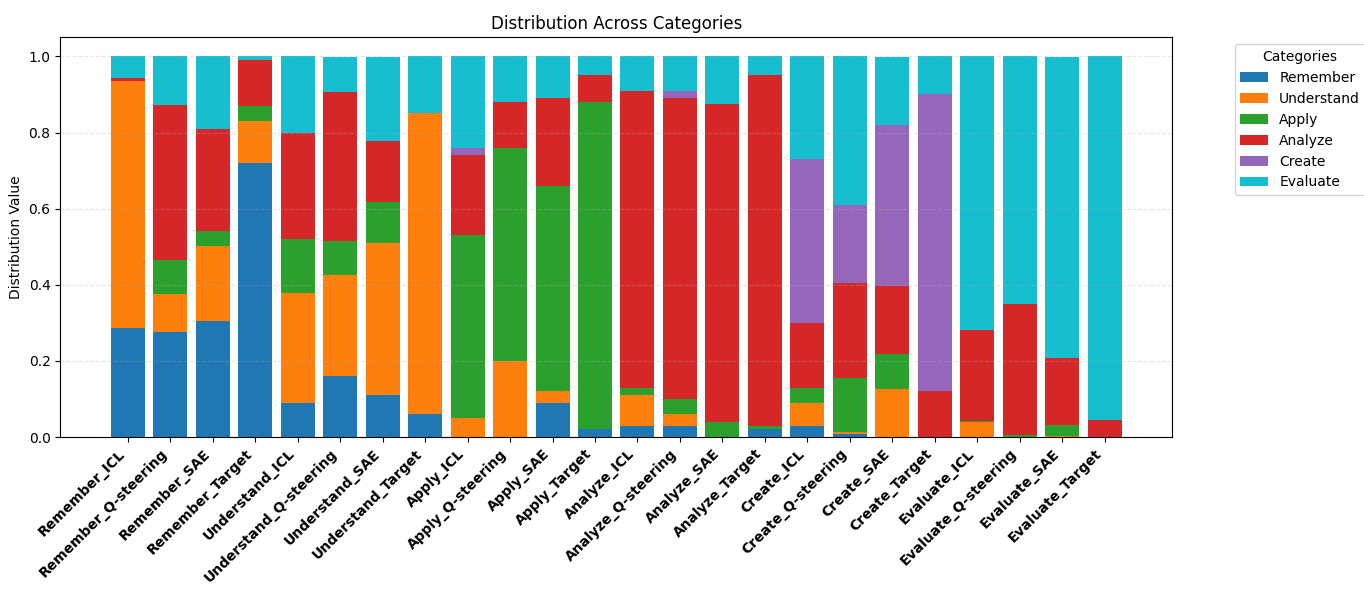}
  \caption{Measured class distribution showing steering accuracy from fine tuned base model output to target cognitive categories. Target classes appear on X-axis while Y-axis shows class category distribution of steered text.}
  \label{fig:scratch}
\end{figure*}

Figure~\ref{fig:scratch} shows the distribution of steered generation category across all the target classes.

\label{sec:LLMoutput}
Below, we present Qualitative examples of the generated responses after steering from fine tuned base model output.

\medskip
\subsubsection*{Steering to apply}
\textbf{Query:}

Below is an instruction that describes a task, paired with an input that provides further context. Write a response that appropriately completes the request.\par
\textbf{Instruction:}\par
Provide feedback on the following code\par
Write a program to increment a number by 1 using a function with call by reference. \par
\textbf{Input}:\par
Code:
\begin{lstlisting}
#include <iostream>
using namespace std;

void increment(int num) {
    num++;
}

int main() {
    int value = 5;
    cout << "Before increment: " << value << endl;
    increment(value);
    cout << "After increment: " << value << endl;
    return 0;
}
\end{lstlisting}

\textbf{Base generation:} The code structure reveals a disconnect between the function's intended purpose and its actual implementation. The increment function is designed to modify a value, but the current implementation fails to achieve this goal due to the missing reference parameter.

\textbf{Steered Generation:} While the code structure includes a function definition and a main function, the call by reference pattern is incorrect. The function needs to be defined with the reference operator to modify the value correctly.

\medskip
\subsubsection*{Steering to create}
\textbf{Query:}\par

Below is an instruction that describes a task, paired with an input that provides further context. Write a response that appropriately completes the request.\par

\textbf{Instruction:}\par
Provide feedback on the following code\par
Write a program to swap two numbers using a function with call by reference.\par

\textbf{Code:}
\begin{lstlisting}
#include <iostream>
using namespace std;

void swap(int a, int b) {
    int temp = a;
    a = b;
    b = temp;
}

int main() {
    int x = 5, y = 10;
    cout << "Before swap: x = " << x << ", y = " << y << endl;
    swap(x, y);
    cout << "After swap: x = " << x << ", y = " << y << endl;
    return 0;
}
\end{lstlisting}

\textbf{Base generation:}  The code demonstrates fundamental pointer concepts with call by reference parameters. However, it's missing the core swap operation implementation where the values need to be exchanged rather than the references being swapped.

\textbf{Steered Generation:} Consider enhancing the solution by implementing proper reference variables, either using pointers or implementing it as a class with member references, or explore in-place swapping techniques using reference operators for a more elegant solution.

\medskip

\subsubsection*{Steering to remember}

\textbf{Query:}\par
Below is an instruction that describes a task, paired with an input that provides further context. Write a response that appropriately completes the request.
\par

\textbf{Instruction:}\par
Provide feedback on the following code\par
Write a program to double all elements of an array using a function with call by reference.
\par

\textbf{Code:}
\begin{lstlisting}
#include <iostream>
using namespace std;

void doubleElements(int arr[5]) {
    for(int i = 0; i < 5; i++) {
        arr[i] = arr[i] * 2;
    }
    return arr;
}

int main() {
    int arr[5] = {1, 2, 3, 4, 5};
    cout << "Before: ";
    for(int i = 0; i < 5; i++) cout << arr[i] << " ";
    doubleElements(arr);
    cout << "\nAfter: ";
    for(int i = 0; i < 5; i++) cout << arr[i] << " ";
    return 0;
}

\end{lstlisting}

{\textbf{Base generation:} The code demonstrates basic array manipulation using a for loop and showcases fundamental function structure with a parameter array. However, the return statement is missing from the main function, which is essential for pointer-based array handling. }

\textbf{Steered Generation:} The relationship between arrays and function parameters shows good comprehension of how arrays can be passed around as values through pointers. The concept that arrays decay to pointers when passed to functions is well-illustrated in your code.

\section{Assigning target center to the decoder}
\begin{figure}[!ht]
\centering
\includegraphics[width=\linewidth]{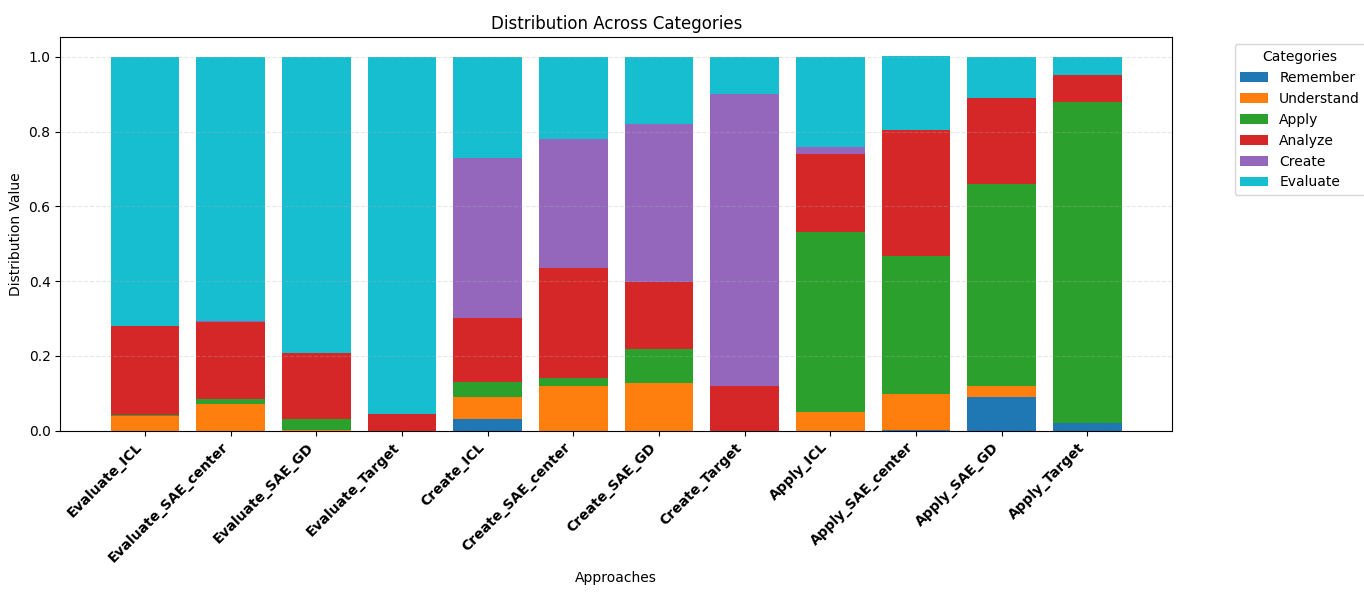}
\caption{Measured class distribution showing steering accuracy. Target classes appear on X-axis while Y-axis shows class category distribution of steered text.}
\label{fig:ltarget center}
\end{figure}

We evaluate the direct use of target center embedding for steered generation (Figure~\ref{fig:ltarget center}). We observe instead of rigid assignment of target center embedding, gradient descent based approach captures 
the target cognitive style effectively. 
\label{sec:tcd}

\clearpage